\documentclass{article}
\usepackage{amsmath,graphicx,mlspconf}

\usepackage{cite}
\usepackage{amsmath,amssymb,amsfonts,mathtools}
\usepackage{bm}
\usepackage{algorithmic}
\usepackage{textcomp}
\usepackage{xcolor}
\usepackage{bbm}
\usepackage{enumitem}
\usepackage{algorithmic}
\usepackage{multirow}
\usepackage{enumitem}
\setlist[enumerate]{topsep=0.5ex,itemsep=0ex,partopsep=1ex,parsep=0.5ex}
\usepackage[margin=0.3em]{subfig}

\usepackage[linesnumbered,ruled,vlined]{algorithm2e}

\newcommand{\argmin}{\mathop{\arg\min}}

\def\BibTeX{{\rm B\kern-.05em{\sc i\kern-.025em b}\kern-.08em
    T\kern-.1667em\lower.7ex\hbox{E}\kern-.125emX}}

\title{Multiscale Graph Construction Using 
Non-local Cluster Features
}

\name{%
   Reina Kaneko$^{\star}$%
   \quad Hayate Kojima$^{\dagger}$%
   \quad Kenta Yanagiya$^{\star}$%
   \quad Junya Hara$^{\star}$%
   \quad Hiroshi Higashi$^{\star}$%
   \quad Yuichi Tanaka$^{\star}$
   \thanks{This work is supported in part by JST AdCORP under Grant JPMJKB2307 and JSPS KAKENHI under Grant 23K17461.}
}
\address{%
   $^{\star}$ Osaka University, Japan\\
   $^{\dagger}$ Tokyo University of Agriculture and Technology, Japan%
}
\begin{document}
\ninept

\maketitle

\begin{abstract}
This paper presents a multiscale graph construction method using both graph and signal features.
Multiscale graph is a hierarchical representation of the graph, where a node at each level indicates a cluster in a finer resolution.
To obtain the hierarchical clusters, existing methods often use graph clustering; however, they may ignore signal variations. As a result, these methods could fail to detect the clusters having similar features on nodes. 
In this paper, we consider graph and node-wise features simultaneously for multiscale clustering of a graph.
With given clusters of the graph, the clusters are merged hierarchically in three steps:
1) Feature vectors in the clusters are extracted.
2) Similarities among cluster features are calculated using optimal transport.
3) A variable $k$-nearest neighbor graph (V$k$NNG) is constructed
and graph spectral clustering is applied to the V$k$NNG to obtain clusters at a coarser scale.
Additionally, the multiscale graph in this paper has \textit{non-local} characteristics: Nodes with similar features are merged even if they are spatially separated.
In experiments on multiscale image and point cloud segmentation, we demonstrate the effectiveness of the proposed method.

\end{abstract}


\section{Introduction}
\label{sec:intro}
Graph signal processing (GSP) is a powerful tool for analyzing signals distributed irregularly in space \cite{ortega_graph_2018}.
Graph signals are defined as signals on a network.
Examples of graph signals are data measured by physical and physiological sensors, LiDAR, and radar, to name a few \cite{wu_graphbased_2017,wagner_distributed_2005}.

In many cases, a graph often forms several clusters and each cluster has nodes having similar features, and therefore, graph signals exhibit a \textit{piecewise smooth} variation in the clusters.
For example, physical sensor networks typically form clusters because of the discontinuity and spatial distance of the measured environment.
This leads to a requirement for analyzing graph signals with several spatial resolutions to detect clusters correctly.

\textit{Multiscale graph} is useful for representing a graph with several spatial resolutions \cite{cour_spectral_2005}. A multiscale graph has a hierarchical structure where a node in a coarser scale graph corresponds to a cluster in a finer one.
It is needed in many GSP applications, including multiresolution analysis of graphs, graph filter bank designs, and graph neural networks \cite{devicofallani_graph_2014,tremblay_subgraphbased_2016}.

\begin{figure}[t]
    \centering
    \subfloat[Conventional multiscale graph.]{\includegraphics[width = .35 \linewidth]{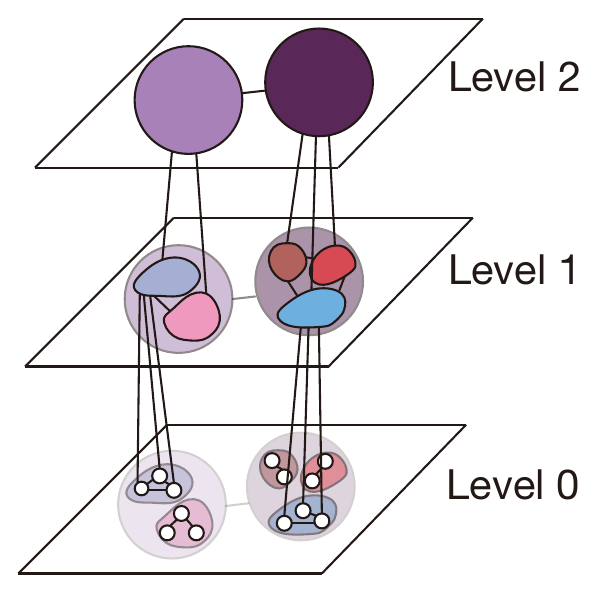}}
    \hspace{1em}
    \subfloat[Proposed multiscale graph.]{\includegraphics[width = .35 \linewidth]{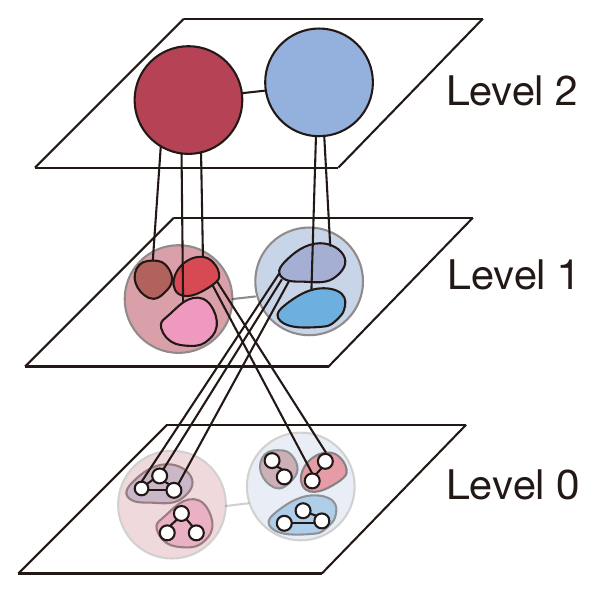}}
    \caption{Comparison of multiscale graphs. Our method can connect non-local clusters, while the conventional one cannot do so.
    }
    \label{multiscalegraph}
\end{figure}

To create a multiscale graph from a given graph, existing methods usually use spectral clustering of nodes in each level \cite{azran_spectral_2006}. It estimates clusters so that the sum of the \textit{cut} (total edge weights among clusters) is minimized based on some criteria.
However, these methods ignore \textit{signal variations} of the nodes/clusters. 
As a result, they could not detect cluster boundaries associated with signal variations appropriately.
Furthermore, as shown in Fig. \ref{multiscalegraph}, existing methods may not connect \textit{non-local} clusters because these methods only consider spatial relationships.
This implies that they cannot structurally detect semantically similar clusters.

To tackle the above-mentioned issues, in this paper, we propose a multiscale graph clustering method by considering graph and node-wise features simultaneously.
Our method takes into account not only the spatial characteristics of the graph but also the signal variation for the multiscale graph construction.

For given clusters of the graph at the finest resolution, we hierarchically merge the clusters in three steps: 
1) $K$-means clustering of features in each cluster is performed to obtain sub-clusters, and their $K$ centroids are extracted as representative feature vectors in the corresponding clusters. 
2) Similarities/proximities among clusters are evaluated based on optimal transport \cite{villani_optimal_2009}.
3) We construct a 
variable $k$-nearest neighbor graph (V$k$NNG) \cite{tamaru_optimizing_2024} based on the similarities among clusters.
Finally, graph spectral clustering \cite{luxburg_tutorial_2007} is applied to the V$k$NNG and obtains coarser-level clusters. 
As illustrated in Fig. \ref{multiscalegraph}, our multiscale graph construction can connect non-local clusters when their features are similar to each other.

In experiments on multiscale image and point cloud segmentation, the proposed method exhibits a comparable performance to existing single-scale methods specifically designed for image processing while we preserve the boundaries of the original clusters in the finest level and connect non-local regions.

\textit{Notation:} We summarize the important notations used throughout the paper in Table~\ref{tab:my_label}.
\section{Related Work}
\label{sec:related}
There are various studies on graph clustering and image/point cloud segmentation.
In this section, first, we introduce graph clustering methods, and then
we review representative image and point cloud segmentation methods.

\subsection{Graph Clustering}
Various graph clustering methods have been proposed so far \cite{hartigan_algorithm_1979,day_efficient_1984} and they are mainly categorized into two approaches: Single-scale and hierarchical multiscale clustering.

    Single-scale clustering is the majority in this field.
    The $K$-means method \cite{hartigan_algorithm_1979} is the most famous one and it partitions data into a fixed number of clusters (i.e., $K$) by iteratively assigning data points to the nearest centroid and updating centroids based on the mean of assigned points. However, since it is designed for the single-scale, revealing a multiscale relationship is usually difficult.
    
    To tackle this issue, a hierarchical clustering method is proposed in \cite{day_efficient_1984}.
    It merges similar clusters in a hierarchical manner.
    Since this method uses all data points (i.e., no subsampling is performed), its computational cost will be high.

\subsection{Image Segmentation}
Graph- and deep learning-based approaches for image segmentation are related to the proposed method.
    
    In \cite{cour_spectral_2005}, a multiresolution pyramid is built from an image and multiscale graphs are obtained by connecting adjacent pixels at each resolution. 
    It imposes constraints to ensure the preservation of object boundaries across both fine and coarse resolutions, and performs spectral clustering across all levels to achieve segmentation. 
    However, since the image pyramid is constructed based on uniform subsampling, the boundaries of smaller objects tend to be less preserved at coarser resolutions. 
    
    To address this, a graph-based multiscale image segmentation method has been proposed \cite{zhang_image_2021}. 
    This is a top-down approach where the input image is partitioned into blocks and they are divided repeatedly until the variation in features within the block is below the user-specified threshold.
    This results in a fine clustering around small object boundaries.
    However, top-down approaches often ignore ``soft'' boundaries like gradations in images.
    
    Deep learning-based methods are widely used for image segmentation. They can be further categorized into two types: Methods using fully convolutional networks (FCNs) \cite{long_fully_2015} and autoencoders \cite{bank_autoencoders_2020}. FCN differs from conventional convolutional neural networks
    as they are composed solely of regular and transposed convolutional layers.
    The transposed layers are shallower than the regular ones, resulting in an asymmetrical structure.
    However, this may lose the clustering boundaries. To overcome this challenge, the autoencoder-based approaches like SegNet \cite{badrinarayanan_segnet_2017} and U-Net \cite{ronneberger_unet_2015} are proposed.
    However, they cannot be straightforwardly used in other data types like point clouds.

\subsection{Point Cloud Segmentation}
\label{related:pointcloud}
Point clouds have irregular and heterogeneous spatial structures, unlike images, complicating the use of multiresolution representations.
Two types of point cloud segmentation approaches, graph- and deep learning-based ones,
have mainly been proposed.

     The $k$-nearest neighbor ($k$NN) method \cite{cover_nearest_1967} is a typical graph-based approach for point cloud clustering.
    Conventional spectral clustering metrics (such as ratio cut) tend to equalize the density of clusters, however, it often results in the violation of boundaries for smaller clusters. A method employing modularity, a metric that enhances the density of edges within each cluster, has also been proposed \cite{tsitsulin_graph_2020}.
    However, this approach relies on a single-scale analysis.
    Therefore, multiscale clustering cannot be performed.
    
    Deep learning-based approaches are often used for point cloud segmentation.
    In \cite{qi_pointnet_2017}, PointNet
    is adapted to learn local features to facilitate point cloud segmentation. Specifically, this involves an iterative process where data points with similar features
     are integrated and re-fed into PointNet.
    Each PointNet block is connected after uniform subsampling and merging $k$-nearest neighbor nodes. However, this approach may lose the boundaries of smaller objects similar to the image pyramid-based analysis.

\vspace{0.1in}
In summary, despite aiming for the (partly) same objectives, distinct approaches have been proposed for different fields. 
In these methods, the following limitations arises:
\begin{enumerate}
    \item Limited datatypes.
    \item Removal of soft boundaries.
    \item Non-preserved boundaries especially for small objects.
\end{enumerate}
In the following, we propose a generic multiscale graph clustering method.

\begin{table}[t]
    \centering
    \caption{Notations used in this paper.}
    \begin{tabular}{c|c}
        \hline
        $\mathbf{Z}^{(l)}$ & Distance matrix at the $l$th scale\\
        $\mathbf{A}^{(l)}$ & Adjacency matrix at the $l$th scale\\
        $L$ & Number of scales of the multiscale graph\\
        $N^{(l)}$ & Number of clusters at the $l$th scale\\
        $\{S_{j}^{(l)}\}_{j = 1,\ldots,N_l}$ & Clusters in the $l$th scale\\
        $\mathcal{F}_{j}^{(l)}$ & Set of centroid feature vectors in $S_{j}^{(l)}$\\
        $\{C_{j,p}\}_{p = 1,\ldots,K_j}$ & Set of sub-clusters in the cluster $S_{j}^{(l)}$\\
        $K_j$ & Number of sub-clusters in the $j$th cluster\\
        ${\mathbf{f}}_{p}^\star$ & Representative feature vector in $C_{j,p}$\\\hline
        $[\mathbf{Z}]_{i,j}$ and $[\mathbf{Z}]_{:i}$ & $(i,j)$th element and $i$th column of $\mathbf{Z}$\\
        $|S|$ & Cardinality of the set $S$\\\hline
    \end{tabular}
    \label{tab:my_label}
\end{table}

\section{Preliminaries}
In this section, we briefly introduce the building blocks used in our proposed method.

\subsection{Optimal Transport}
Optimal transport \cite{villani_optimal_2009} is a method for comparing two distinct feature spaces. It calculates the costs of transportation beforehand, allowing it to measure the distance between data from structurally different feature spaces.

Let $X_{j_1}$ and $X_{j_2}$ be two different feature spaces under consideration indexed by $j_1$ and $j_2$, and let $\bm{\mu}_{j_1}\in\mathbb{R}_{>0}^{|X_{j_1}|}$ and $\bm{\mu}_{j_2}\in\mathbb{R}_{>0}^{|X_{j_2}|}$ be their Borel measures. Optimal transport between $\bm{\mu}_{j_1}$ and $\bm{\mu}_{j_2}$ is described as follows:
\begin{equation}
        \begin{gathered}
        [\mathbf{Z}]_{j_1,j_2}\coloneqq \text{OT}(X_{j_1},X_{j_2})= \underset{\mathbf{P}}{\min} \hspace{-0.1in}\sum_{(p_1,p_2)\in X_{j_1}\times X_{j_2}}\hspace{-0.1in}[\mathbf{P}]_{p_1,p_2}[\mathbf{C}]_{p_1,p_2}\\
        \text{s.t. }\mathbf{P}\mathbf{1}=\bm{\mu}_{j_1},\,
        \mathbf{P}^\top\mathbf{1} = \bm{\mu}_{j_2},\,[\mathbf{P}]_{p_1,p_2}\geq0,\\
\end{gathered}\label{eq:ot}
\end{equation}
where $\mathbf{P}$ and $\mathbf{C}$ are the transportation and cost matrices between $X_{j_1}$ and $X_{j_2}$, respectively. 
We can solve \eqref{eq:ot} by standard linear programming.

\subsection{V$k$NNG}
The $k$NN\cite{cover_nearest_1967} is one of the most popular approaches for graph construction and is widely used in machine learning and signal processing. 
An improved version of the $k$NNG, called V$k$NNG, is proposed in \cite{tamaru_optimizing_2024}.

Let $\mathbf{Z}\in\mathbb{R}_{\geq 0}^{N\times N}$ be a given distance matrix, where $[\mathbf{Z}]_{i,j}$ indicates the distance between nodes $i$ and $j$. It optimizes $k$ for each node with the following objective function:
\begin{equation}
    \mathbf{a}_{\text{variable},k_i}\coloneqq\argmin_{\mathbf{a}_i\in\{0,1\}^{N}}\|\mathbf{a}_i\circ[\mathbf{Z}]_{i,:}\|_1-\alpha_i\|\mathbf{a}_i\|_1,\label{eq:tamaru}
\end{equation}
where $(\circ)$ is the Harmard product and $\alpha_i$ is a properly chosen parameter (see \cite[Sec.3.3]{tamaru_optimizing_2024} for more details). By solving \eqref{eq:tamaru} for each node, an \textit{unweighted} adjacency matrix $\mathbf{A}_{\text{uw}}=[\mathbf{a}_{\text{varied},k_1},\ldots,\mathbf{a}_{\text{varied},k_{N}}]\in \{0,1\}^{N\times N}$ of the V$k$NNG is obtained. 
In this paper, we use its weighted version given by $\mathbf{A}=\mathbf{A}_{\text{uw}}\circ \mathbf{Z}$.

\section{Multiscale Graph Clustering Method} %
\label{sec:methods}
In this section, we introduce the proposed multiscale graph construction based on optimal transport and V$k$NNG.
\subsection{Overview}
\begin{figure}[t]
\centering
\includegraphics[width=0.7\linewidth]{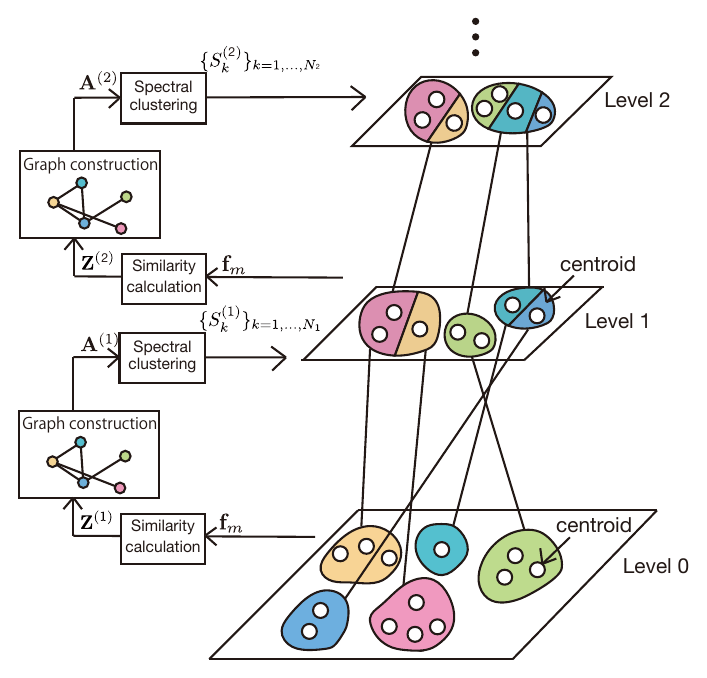}
\caption{Overview of multiscale graph signal clustering.}
\label{graph}
\end{figure}

Let $G=(V,E)$ be an initial graph, where $V$ and $E$ are the sets of nodes and edges.
Hereafter, $\cdot^{(l)}$ represents entities in the $l$th scale.
Let $S_j^{(l)}\subset V$ ($l = 0, \dots, L$) be the $j$th cluster of nodes at the $l$th scale where smaller $l$ corresponds to a finer scale. We denote the number of clusters at the $l$th level by $N^{(l)}$.

Here, we assume that $G^{(0)} = G$ and the clusters at the finest resolution $\{S_j^{(0)}\}_{j=1,\ldots,N^{(0)}}$ are given. We hierarchically merge the clusters to obtain a multiscale graph.
Therefore, the segmentation boundaries in the finest scale are preserved.
Fig.~\ref{graph} illustrates the overview of the proposed method.

\begin{figure}[t]
\centering
\includegraphics[width=0.9\linewidth]{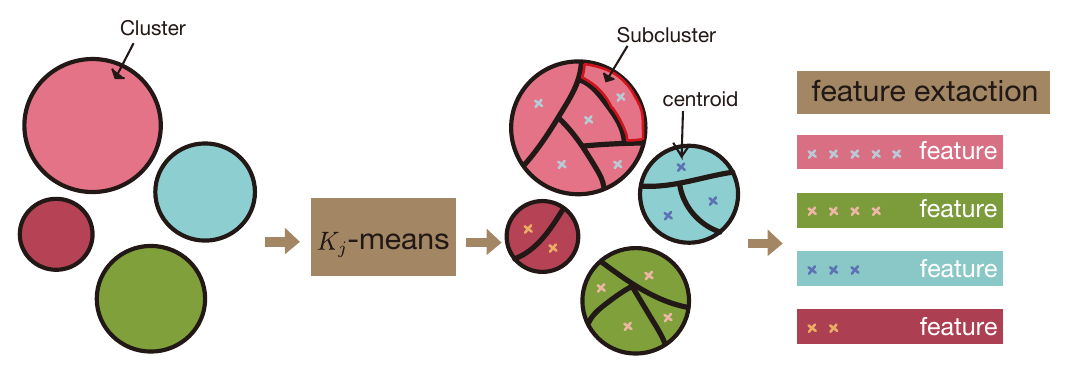}
\caption{Overview of feature extraction.}
\label{fig:feature_ext}
\end{figure}

\begin{algorithm}[t]
 \caption{Multiscale clustering from the $l$th scale to the $(l+1)$ th scale}
 \label{al:spectral}
 \begin{algorithmic}[1]
 \renewcommand{\algorithmicrequire}{\textbf{Input:}}
 \renewcommand{\algorithmicensure}{\textbf{Output:}}
 \REQUIRE $G^{(l)}$, $\{S_j^{(l)}\}_{j=1,\ldots,N^{(l)}}$, $N^{(l)}$, and $N^{(l+1)}$
 \ENSURE  $G^{(l+1)}$, $S_1^{(l+1)}, \ldots, S_{N^{(l+1)}}^{(l+1)}$
 \\
 \textit{Feature extraction}\\
  \FOR {each clusters in $S_1^{(l)}, \ldots, S_{N^{(l)}}^{(l)}$}
  \STATE $\mathbf{f}_{(j,p)}^\star$ $\gets$ feature vector of centroid in $C_{j,p}^\star$ in \eqref{eq:k_j_means}
  \ENDFOR
  \\
  \textit{Calculation of distances}\\
  \FOR {each combination of two clusters in $(S_{j_1}^{(l)},S_{j_2}^{(l)})$}
  \STATE  $[\mathbf{Z}^{(l+1)}]_{j_1,j_2}$ $\gets$ distance between $\mathcal{F}_{j_1}^{(l)}$ and $\mathcal{F}_{j_1}^{(l)}$   in \eqref{eq:ot}
  \ENDFOR
  \\
  \textit{Graph construction}\\
  \STATE $\mathbf{A}^{(l+1)}$ $\gets$ adjacency matrix of $G^{(l+1)}$ in \eqref{eq:tamaru}
  \\
  \textit{Spectral clustering}\\
  \STATE $\{S_1^{(l+1)}\}_{j=1,\ldots,N^{(l+1)}}$ $\gets$ clusters obtained by $\text{Normalized-cut}(\mathbf{A}^{(l+1)},N^{(l+1)})$\\
  \STATE $l$ $\gets$ $l+1$
 \end{algorithmic} 
 \end{algorithm}

\begin{figure*}[t]
    \centering
    \subfloat[Original image and segment boundaries of SLIC (yellow lines)]{\includegraphics[width = .25 \linewidth]{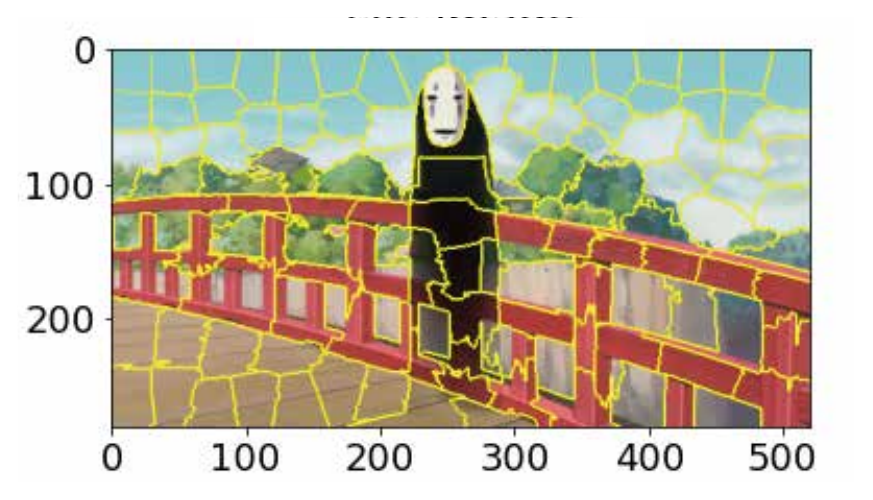}}
    \subfloat[Segmented image by our method (OT)]{\includegraphics[width = .25 \linewidth]{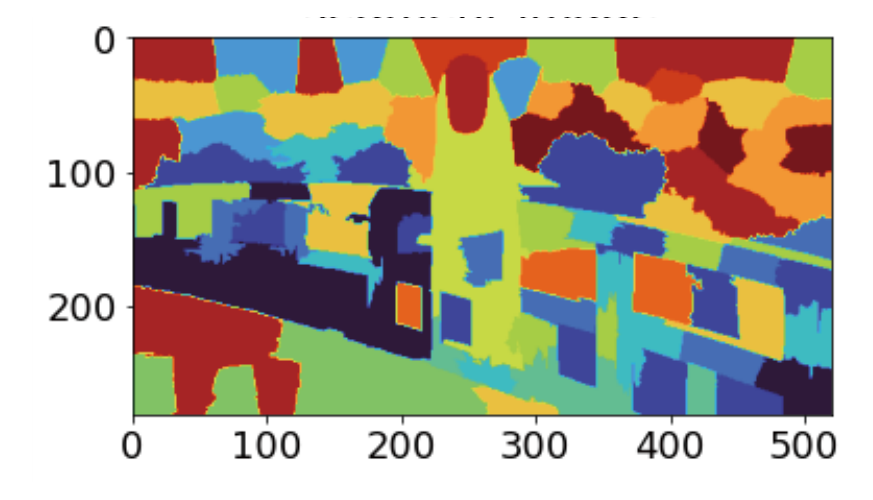}}
    \subfloat[Segmented image by SLIC]{\includegraphics[width = .25 \linewidth]{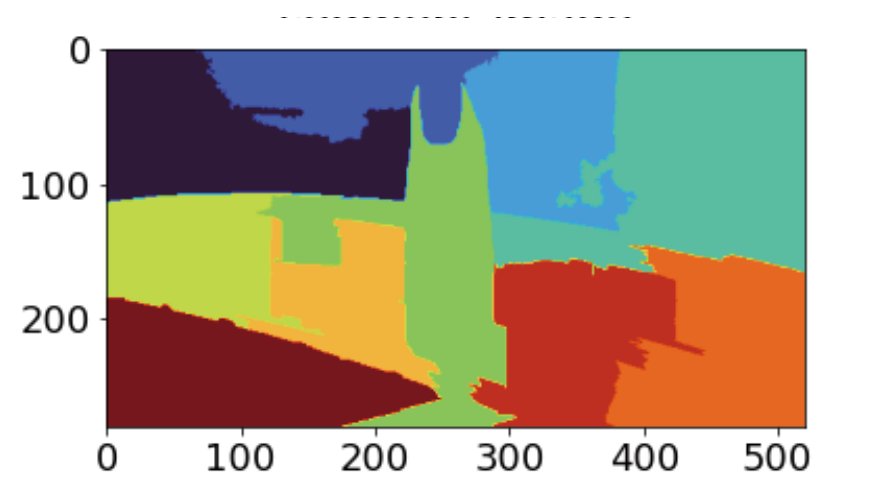}}
    \subfloat[Segmented image by MNCut]{\includegraphics[width = .25 \linewidth]{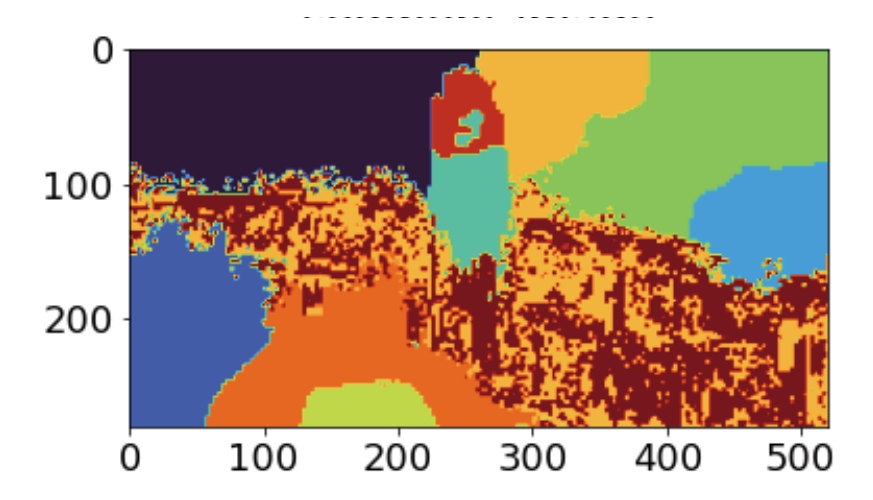}}\\\vspace{-0.2in}
    \subfloat[Segmented image by semantic segmentation]{\includegraphics[width = .25 \linewidth]{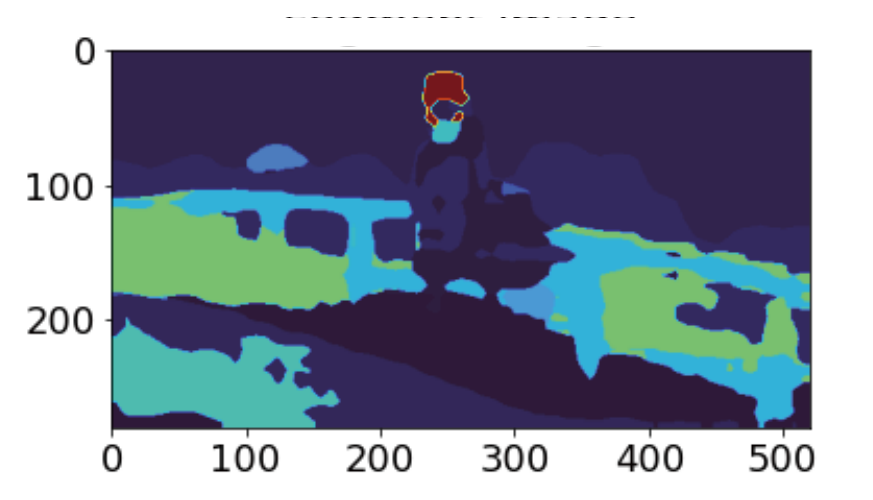}}
    \subfloat[Segmented image by our method (OT)]{\includegraphics[width = .25 \linewidth]{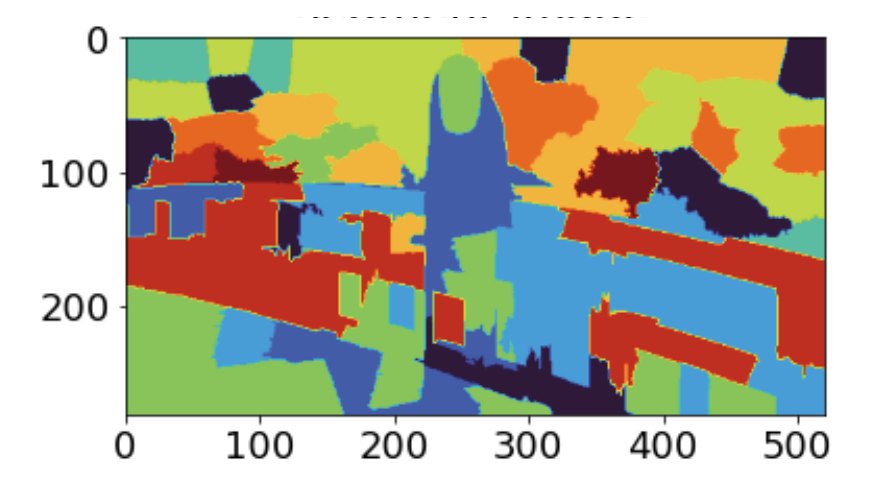}}
    \subfloat[Segmented image by SLIC]{\includegraphics[width = .25 \linewidth]{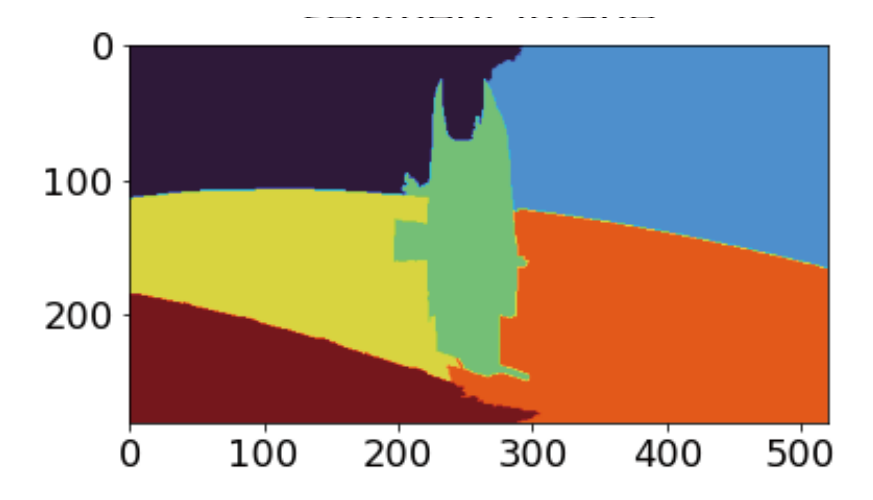}}
    \subfloat[Segmented image by MNCut]{\includegraphics[width = .25 \linewidth]{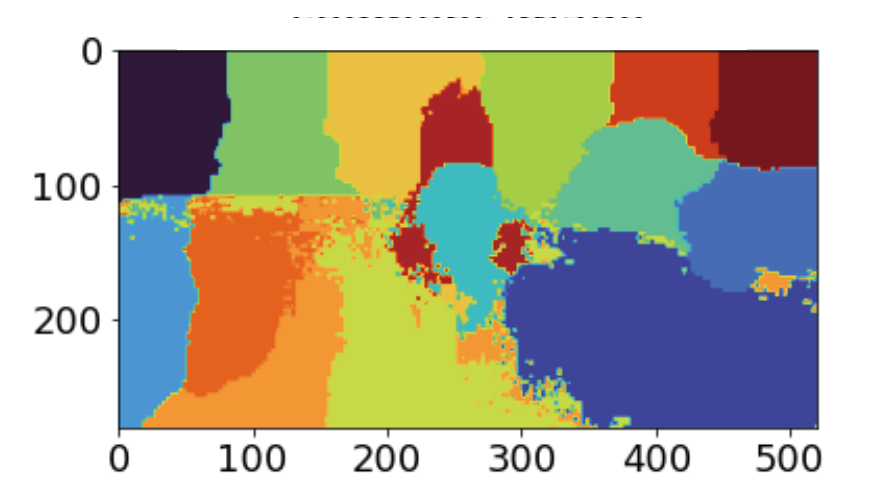}}\\
    \caption{Image segmentation results. Top: 15 clusters. Bottom: 10 clusters.
    }
    \label{seg_img_result}
\end{figure*}

In the following, we present our algorithm from the $l$th scale to the $(l+1)$th scale.
We merge the clusters in three steps:
\begin{enumerate}
    \item Feature extraction.
    \item Calculation of similarities among $S_j^{(l)}$.
    \item Graph construction and spectral clustering.
\end{enumerate}
The above process is repeated from $l=0$ until $l$ reaches $L$.
We describe the details of the steps below. We show the algorithm for the $l$th scale in Algorithm \ref{al:spectral}.

\subsection{Feature Extraction}
\label{sec:feature}
Suppose that $\mathbf{f}_{m}\in\mathbb{R}^F$ is a feature vector of the $m$th node.
To alleviate the computational burden using all feature vectors in $S_j^{(l)}$ for all $j$ ($j=1,\dots, N^{(l)}$) in the subsequent steps, we subsample the representative features with the following steps.
We also illustrate the feature extraction step in Fig. \ref{fig:feature_ext}.

\begin{enumerate}
    \item We first divide $S_j^{(l)}$ into sub-clusters $\{C_{j,p}^\star\}_{p=1,\dots,K_j}$ by $K_j$-means clustering, where the number of sub-clusters $K_j$ is given\footnote{For notation simplicity, we omit the superscript $\cdot^{(l)}$ for $C_{j,p}^\star$ and $K_j$.}.
    The $K_j$-means clustering is formulated as 
    \begin{equation}
        \{C_{j,p}^\star\}_{p=1,\ldots,K_j}\coloneqq\argmin_{\{C_{j,p}\}_{p=1,\ldots,K_j}}\sum_{p=1}^{K_j}\sum_{m\in C_{j,p}}\|\mathbf{f}_{m}-\bar{\mathbf{f}}_p\|^2,\label{eq:k_j_means}
    \end{equation}
    where $\bar{\mathbf{f}}_p=\frac{1}{|C_{j,p}|}\sum_{{m}\in C_{j,p}} \mathbf{f}_{m}$ is the centroid vector in $C_{j,p}$. 
    \item 
    We obtain one centroid node from $C_{j,p}^\star$ and set the associated feature vector
    as the representative sub-cluster features, i.e., $\mathbf{f}^\star_p=\bar{\mathbf{f}}_p$.
    Accordingly, a set of representative vectors $\mathcal{F}_{j}^{(l)}=\{\mathbf{f}_p^\star\}_{p=1, \dots, K_j}$ is given as the features in the cluster $S_j^{(l)}$.
    Since we can include signal values in $\mathbf{f}_{p}^\star$, we can utilize signal variations as well as spatial features.
\end{enumerate}

\subsection{Calculation of Similarities Among $\{S_j^{(l)}\}$}
Here, we consider two sets of features $\mathcal{F}_{j_1}^{(l)}$ and $\mathcal{F}_{j_2}^{(l)}$ $(j_1, j_2 \in \{1, \dots, N^{(l)}\})$ associated with the two clusters $S_{j_1}^{(l)}$ and $S_{j_2}^{(l)}$.
To obtain their similarity, we solve $\text{OT}(\mathcal{F}^{(l)}_{j_1},\mathcal{F}^{(l)}_{j_2})$ in \eqref{eq:ot}, in which the cost matrix $[\mathbf{C}]_{p_1,p_2}$ in \eqref{eq:ot} is given by
\begin{equation}
    [\mathbf{C}]_{p_1,p_2}=\|\mathbf{f}_{p_1}^\star-\mathbf{f}_{p_2}^\star\|^2 \quad \forall(p_1,p_2)\in \mathcal{F}_{j_1}^{(l)}\times \mathcal{F}_{j_2}^{(l)},\label{eq:cost_matrix}
\end{equation}
and $\bm{\mu}_{j_1},\bm{\mu}_{j_2}$ in \eqref{eq:ot} are defined as
\begin{equation}
\begin{aligned}
    \bm\mu_{j_1}[p_1]\coloneqq& |C_{j_1,p_1}^\star|/|S_{j_1}| \quad p_1=1,\ldots,K_{j_1},\\
    \bm\mu_{j_2}[p_2]\coloneqq& |C_{j_2,p_2}^\star|/|S_{j_2}| \quad p_2=1,\ldots,K_{j_2}.
\end{aligned}
\end{equation}
Consequently, we have the similarity matrix $\mathbf{Z}^{(l+1)}$ by solving \eqref{eq:ot} for all $(j_1,j_2)\in \{1,\ldots,N^{(l)}\}\times \{1,\ldots,N^{(l)}\}$.

\subsection{Graph Construction and Spectral Clustering}
We construct a V$k$NNG from the similarity matrix $\mathbf{Z}^{(l+1)}$ obtained in the previous step. The optimal $k_i$ for $G^{(l+1)}$ is determined by \eqref{eq:tamaru} and the adjacency matrix $\mathbf{A}^{(l+1)}$ for a coarser scale is obtained. Note that we can connect non-local clusters by the use of V$k$NNG.

\begin{figure}[t]
\centering
\includegraphics[width=0.7\linewidth]{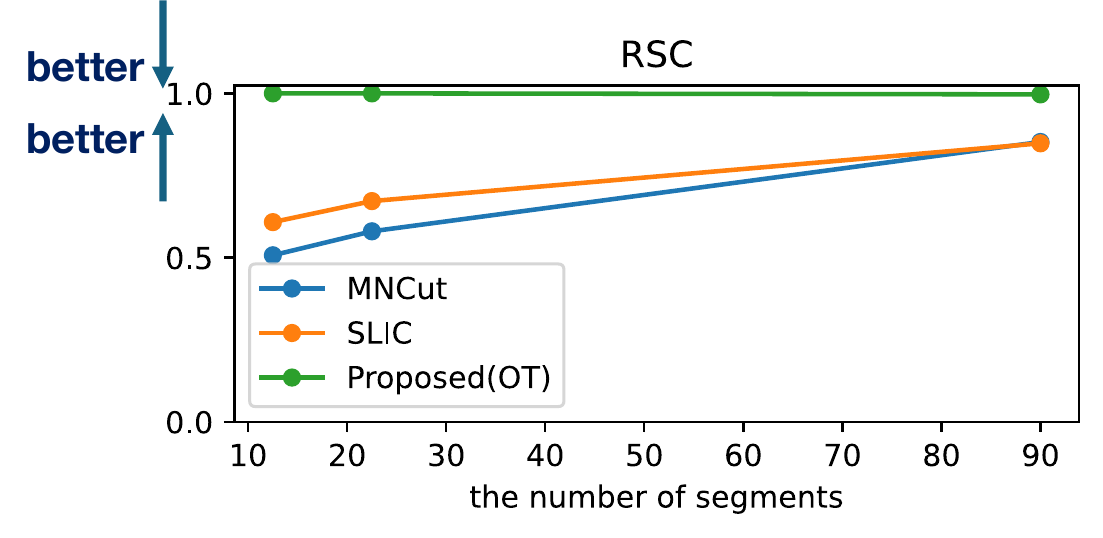}
\caption{Mean of RSC for all clusters for image segmentation. We show the averaged values for all three images.
}
\label{fig:imgiou}
\end{figure}

Finally, we obtain coarser-level clusters from $\mathbf{A}^{(l)}$ by solving the normalized cut problem (see Sec.~\ref{related:pointcloud}).

\section{Experimental results}
\label{sec:results}
In this section, we perform experiments on image and point cloud segmentation using our method and their results are compared with existing methods.

\subsection{Image Segmentation}

\begin{table}[t!]
\centering
  \caption{Mean of cluster-wise standard deviation of RGB values. For the proposed method, standard deviations for 10 and 15 clusters correspond to those in the $L$th and $(L-1)$th layers, respectively. SS denotes the semantic segmentation method \cite{zhou_semantic_2019}.}
  \label{img_iou}
\begin{tabular}{|c|c|c|c|c|}
\hline
\multirow{2}{*}{$N^{(l)}$} & \multirow{2}{*}{Ours} & \multirow{2}{*}{MNCut} & \multirow{2}{*}{SLIC} & \multirow{2}{*}{SS} \\
                  &                       &                        &                       &                     \\ \hline
10                & 15.51                 & 15.32                  & 17.97                 & --                  \\ \hline
15                & 12.24                 & 12.60                  & 14.87                 & 19.17               \\ \hline
\end{tabular}
\end{table}

\begin{figure*}[t]
    \centering
    \subfloat[Original point cloud]{\includegraphics[width = .18 \linewidth]{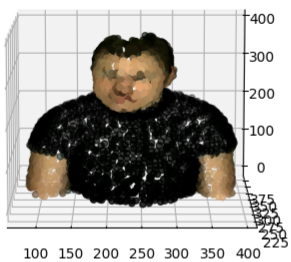}}
    \subfloat[Segmented point cloud by our method (DMoN)]{\includegraphics[width = .18 \linewidth]{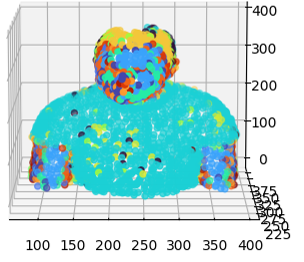}}
    \subfloat[Segmented point cloud by our method ($K$-means)]{\includegraphics[width = .18\linewidth]{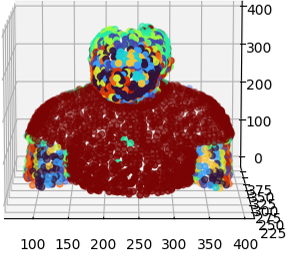}}
    \subfloat[Segmented point cloud by DMoN]{\includegraphics[width = .18 \linewidth]{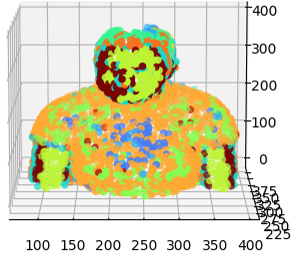}}
    \subfloat[Segmented point cloud by $K$-means]{\includegraphics[width = .18 \linewidth]{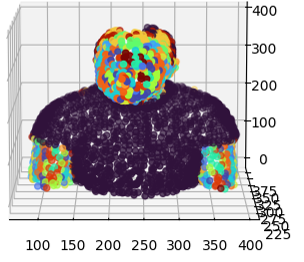}}\\\vspace{-0.15in}
    \subfloat[Segmented point cloud by PointNet++]{\includegraphics[width = .18 \linewidth]{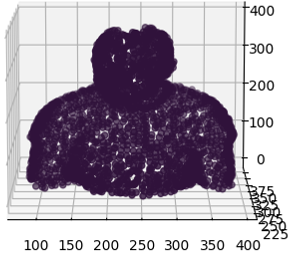}}
    \subfloat[Segmented point cloud by our method (DMoN)]{\includegraphics[width = .18\linewidth]{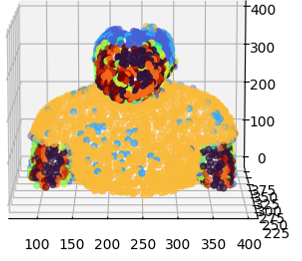}}
    \subfloat[Segmented point cloud by our method ($K$-means)]{\includegraphics[width = .18\linewidth]{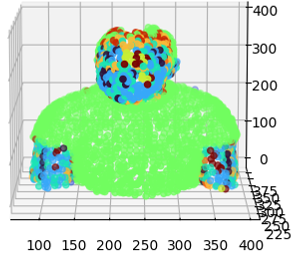}}
    \subfloat[Segmented point cloud by DMoN]{\includegraphics[width = .18 \linewidth]{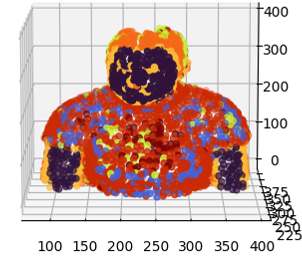}}
    \subfloat[Segmented point cloud by $K$-means]{\includegraphics[width = .18 \linewidth]{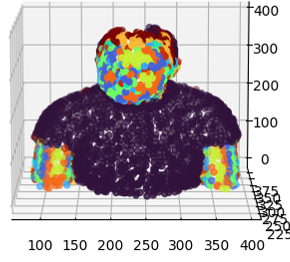}}
    \caption{Point cloud segmentation results. Top: 15 clusters. Bottom: 10 clusters.}
    \label{seg_pcd_result}
\end{figure*}

\begin{figure}[t]
\centering
\includegraphics[width=.7\linewidth]{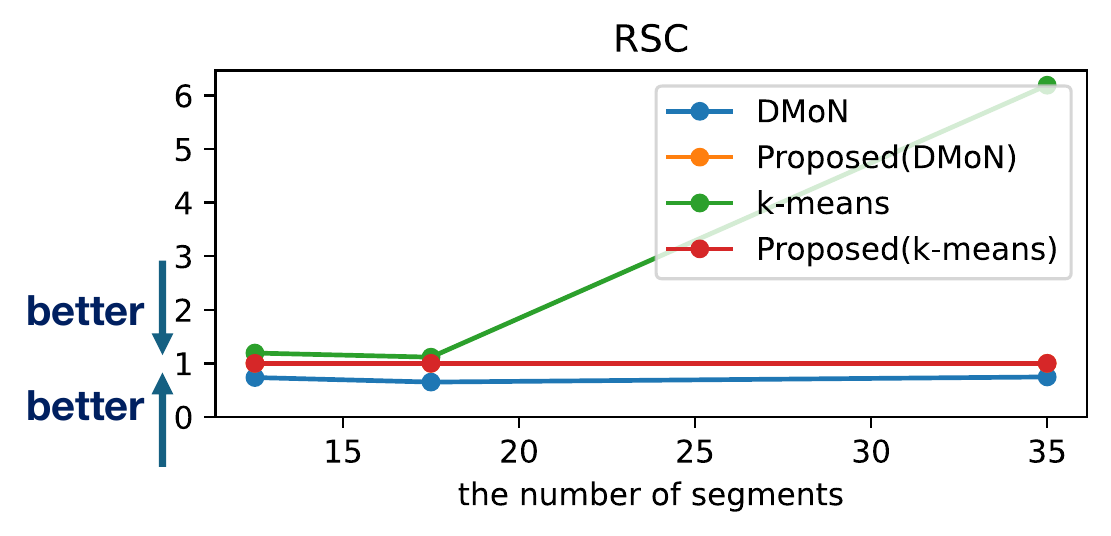}
\caption{Mean of RSC for all clusters for point cloud segmentation. We show the averaged values for all point clouds.}
\label{fig:pcdiou}
\end{figure}

\noindent
\textit{Setup:}
We showcase image segmentation with three images \texttt{No-Face} in \textit{Spirited Away}, \texttt{Ponyo} in \textit{Ponyo on the Cliff by the Sea}, and \texttt{Arrietty} in \textit{Arrietty} taken from movies by STUDIO GHIBLI\footnote{The image is taken from the official website (https://www.ghibli.jp/info/013772/).}.
An example of \texttt{No-Face} is shown in Fig. \ref{seg_img_result}(a).
We set the number of layers to $L=4$ and the number of clusters at the $L$th layer to $N^{(L)}=10$.
For our method, simple linear iterative clustering (SLIC) \cite{khorasgani_slic_2022} is used as the initial image clustering to obtain $\{S_j^{(0)}\}$\footnote{An arbitrary image segmentation method can be applied to our proposed method.}.

We compare our method with the following image clustering methods:
\begin{enumerate}
    \item SLIC
    \item Multiscale normalized cut (MNCut) \cite{cour_spectral_2005}
    \item Deep semantic segmentation \cite{zhou_semantic_2019}
\end{enumerate}

\textit{Results:}
Fig. \ref{seg_img_result} shows the image segmentation results.
For all methods, the same color represents the same cluster.
As observed, the proposed method and the semantic segmentation method connect non-local segments having similar features, while other existing methods cannot do so.
While SLIC yields very large clusters, one can observe that it produces very thin clusters in edge regions.
Furthermore, the proposed method maintains the initial clustering boundaries of the input image as expected, in contrast to the alternative methods. 

We measure the ability for cluster boundary preservation with \textit{ratio of subclusters within a cluster} (RSC):
To begin with, let us consider the following cardinality function between sets $A$ and $B$.
\begin{equation}
    \mathbb{I}(A,B)\coloneqq 
    \begin{cases}
        |A\cup B| & \text{if }A\cap B\neq \emptyset \\
        0 & \text{otherwise.}
    \end{cases}\label{eq:card_func}
\end{equation}
If the target set $A$ is overlapped with the reference one $B$, $\mathbb{I}(A, B)$ counts the number of elements in the union of $A$ and $B$.
With \eqref{eq:card_func}, the RSC score of $S_{i}^{(l+1)}$ is defined as follows:
\begin{equation}
    \text{RSC}(S_{i}^{(l+1)})\coloneqq \frac{1}{|S_{i}^{(l+1)}|}\sum_{j=1}^{N^{(l)}}\mathbb{I}(S_{j}^{(l)},S_{i}^{(l+1)}).\label{eq:RSC}
\end{equation}
Note that \eqref{eq:RSC} evaluates the ratio of the number of nodes between reference cluster and target sub-clusters, which indicates that the closer the ratio is to 1, the more equal the number of nodes between $S_{j}^{(l)}$ and $S_{i}^{(l+1)}$, which results in that the cluster boundary is preserved.

Fig. \ref{fig:imgiou} summarizes the results of the mean of RSC for all clusters.
The proposed method maintains RSC to 1 that demonstrates its ability for cluster boundary preservation.

\begin{table}[t]
\centering
  \caption{Mean of cluster-wise standard deviation of RGB values. For the proposed method, standard deviations for 10 and 15 clusters correspond to those in the $L$th and $(L-1)$th layers, respectively.}
  \label{pcd_std}
\begin{tabular}{|c|cc|c|c|}
\hline
\multirow{2}{*}{$N^{(l)}$} & \multicolumn{2}{c|}{Ours}                                & \multirow{2}{*}{DMoN} & \multirow{2}{*}{$K$-means} \\ \cline{2-3}
                  & \multicolumn{1}{c|}{DMoN} & \multicolumn{1}{c|}{$K$-means} &                       &                          \\ \hline
10                & \multicolumn{1}{c|}{7.47} & 7.56                         & 6.55                  & 6.47                     \\ \hline
15                & \multicolumn{1}{c|}{5.12} & 5.46                         & 4.78                  & 5.22                     \\ \hline
\end{tabular}
\end{table}

Table \ref{img_iou} compares the mean of cluster-wise standard deviations of the RGB pixel values. As observed, our method shows a similar score to existing methods specifically designed for image segmentation.

\subsection{Point Cloud Segmentation}
\noindent
\textit{Setup:}
For point cloud segmentation, we use three point clouds \texttt{Sarah}, \texttt{David} and \texttt{Andrew} taken from \cite{loop_microsoft_2016}.
We set the number of layers to $L=3$ and the number of clusters at the $L$th layer to $N^{(L)}=10$.
We compare our method with three point cloud segmentation methods:
\begin{enumerate}
    \item Deep modularity networks (DMoN) \cite{tsitsulin_graph_2020}
    \item PointNet++ (pretrained with S3DIS) \cite{qi_pointnet_2017}
    \item $K$-means \cite{hartigan_algorithm_1979}
\end{enumerate}
For the proposed method, we utilize the clustering results of DMoN and $K$-means as the input clusters $\{S_j^{(0)}\}$.

\textit{Results:}
Fig. \ref{seg_pcd_result} shows point cloud segmentation results of \texttt{Sarah}.
For all methods, the same color represents the same cluster like in the image segmentation experiment. 
It can be observed that
all methods except for PointNet++ connect similar body parts as the same segments even if they are spatially separated.
The segmentation results by PointNet++ are incorrect due to the inconsistency of the training and test data where it has been trained with the indoor scene dataset \cite{armeni_joint_2017}.
Note that, since our method does not require training data, it can be used for many data types straightforwardly.
Furthermore, the proposed method can maintain the initial cluster boundaries of the input point cloud, in contrast to the alternative methods.

Fig. \ref{fig:pcdiou} shows the mean of RSC for all clusters. Our method demonstrates superior performance compared to the other methods.
Table \ref{pcd_std} shows the mean of the cluster-wise standard deviation of RGB values of the point cloud. The proposed method shows comparable value to the other methods. 

The above-mentioned results imply that the proposed generic graph clustering method is effective for point cloud segmentation as well as image segmentation.

\section{Conclusion}
\label{sec:con}
In this paper, we propose a multiscale graph clustering method using graph and signal features simultaneously. We utilize optimal transport and V$k$NN graphs for realizing the multiscale clustering. We hierarchically merge clusters with the following three steps: 1) Extracting feature vectors from each cluster, 2) calculating similarties using optimal transport, and 3) constructing the graph using V$k$NN and applying spectral clustering.
Our proposed approach can connect non-local similar features and preserve the original boundaries at the finest resolution.
The experimental results in image and point cloud segmentation validate that our proposed method effectively merges clusters having similar features.
Our future work includes acceleration of the algorithm.


\end{document}